\documentclass{article}

\usepackage[final]{neuripsDeepInverse_2020}

\usepackage[utf8]{inputenc} 
\usepackage[T1]{fontenc}    
\usepackage{hyperref}       
\usepackage{url}            
\usepackage{booktabs}       
\usepackage{amsfonts}       
\usepackage{nicefrac}       
\usepackage{microtype}      

\usepackage{fontawesome}
\usepackage{graphicx}
\usepackage{amsmath}
\usepackage{amssymb}
\usepackage{amsfonts}
\usepackage{color}
\usepackage{xspace}

\def\XS{\xspace}
\DeclareMathAlphabet{\mathb}{OML}{cmm}{b}{it}
\def\sbm#1{\ensuremath{\mathb{#1}}}
  \def\xb{{\sbm{x}}\XS}
  \def\yb{{\sbm{y}}\XS}
\def\Fb{{\sbm{F}}\XS}  

\newcommand{\imgur}{\href{https://imgur.com/a/mqB1F80}{\faPlayCircle}}

\title{Denoising Score-Matching for Uncertainty Quantification in Inverse Problems}

\author{%
  Zaccharie Ramzi\thanks{https://www.cosmostat.org/people/zaccharie-ramzi} \\
  CEA (Neurospin and Cosmostat), Inria (Parietal)\\
  Gif-sur-Yvette, France \\
  \texttt{zaccharie.ramzi@inria.fr} \\
   \And
   Benjamin Remy, Fran\c{c}ois Lanusse, Jean-Luc Starck\\
  AIM, CEA, CNRS \\ 
  Universit\'e Paris-Saclay \\ 
  Universit\'e Paris Diderot \\
  Sorbonne Paris Cit\'e \\
  \{benjamin.remy, francois.lanusse, jean-luc.starck\}@cea.fr
  \And
  Philippe Ciuciu\\
  CEA (Neurospin), Inria (Parietal)\\
  Gif-sur-Yvette, France \\
  philippe.ciuciu@cea.fr
}

\begin{document}

\maketitle

\begin{abstract}
Deep neural networks have proven extremely efficient at solving a wide range of inverse problems, but most often the uncertainty on the solution they provide is hard to quantify. In this work, we propose a generic Bayesian framework for solving inverse problems, in which we limit the use of deep neural networks to learning a prior distribution on the signals to recover. We adopt recent denoising score matching techniques to learn this prior from data, and subsequently use it as part of an annealed Hamiltonian Monte-Carlo scheme to sample the full posterior of image inverse problems. We apply this framework to Magnetic Resonance Image (MRI) reconstruction and illustrate how this approach not only yields high quality reconstructions but can also be used to assess the uncertainty on particular features of a reconstructed image.
\end{abstract}

\section{Introduction}

Generative modeling has enjoyed remarkable success in recent years with models such as Generative Adversarial Networks (GANs)~\citep{Goodfellow2014GenerativeNets} reaching extremely high quality results on complex high resolution images~\citep{Karras2020}. 
Yet, GANs are still prone to issues including unstable training and mode collapse, i.e. a lack of diversity in generated images. 
In addition, GANs only provide a convenient way to sample from the learned distribution, they do not give access to the density function itself. 
Another class of commonly used models, Variational Auto-Encoders (VAEs)~\citep{Kingma2014} have also been able to reach high quality sample resolutions~\citep{Vahdat2020}, but can only provide a bound on the density function of the model. 

In some practical applications of these generative models however, the most desirable aspect is not necessarily to be able to sample from the model, or directly have access to its density functions, but have access to gradients of the log density function, i.e. its \textit{score}. 
In particular, in this work we focus on 
the generic problem of performing Bayesian inference to solve an inverse problem (i.e. deconvolution, denoising, inpainting, etc), using a learned generative model as a Bayesian prior. 
Such problems can typically be solved by gradient-based inference methods such as Variational Inference (VI), and Langevin or Hamiltonian Monte-Carlo (HMC), but all these methods have in common that they \textit{only rely on the score of the generative model}. 

Recent work \citep{Lim2020,Song2019} has investigated various approaches to directly train a deep neural network (coined, a \textit{score network}) to estimate the score of a density function instead of the density itself. 
One of the most promising and scalable approach relies on denoising score-matching, where the score network is trained using a denoising loss on an augmented dataset where various amounts of Gaussian noise is added to the data. 
Not only has this approach already been demonstrated to reach convincing results in score estimation, but~\cite{Song2020} for instance demonstrated that these score networks can be used to sample high resolution natural images through Langevin dynamics. 

In this work, we develop the first application of deep denoising score matching to Magnetic Resonance Image (MRI) reconstruction and its use for Uncertainty Quantification (UQ) in the context of imaging inverse problems.

\section{Related Works}


\paragraph{Score Matching}
Score Matching was introduced by~\cite{HyvarinenAAPOHYVARINEN2005}.
It was later revisited in~\cite{Vincent2011} and~\cite{Alain2013} in the context of Deep Learning and Denoising Auto-Encoders (DAE) in particular.
Recently, \cite{Song2019} applied these findings to images showing the potential of this approach combined with sampling techniques. They refined their work in~\cite{Song2020} to take it to high dimension and provide some techniques for hyper-parameter setting.

\paragraph{Plug-and-Play Priors} The idea of using a denoising model as a prior for inverse problems solving was introduced by~\cite{Venkatakrishnan2013Plug-and-PlayReconstruction} who used non-learned denoisers to replace the proximity operator in the ADMM algorithm~\citep{Boyd2011AlternatingMultipliers}. 
More recently, \cite{Meinhardt2017LearningProblems} have made use of learned denoiser networks in Plug-and-Play inverse problems solving.
Annealed HMC sampling can be seen as a fuzzy-version of these approaches with theoretical grounds.

\paragraph{Deep Uncertainty Quantification in Inverse Problems} The idea of using generative models for inverse problems such as data imputation and denoising was for instance suggested in an early VAE work~\citep{Rezende2014}. In this approach, a generative network is first trained on high quality data, and inference is then performed in the latent space of
the model typically using VI. The main advantage comes from the reduced dimensionality of the parameter space. Recent examples of this approach include~\citep{Wu2018,Bohm2019}, and an application to MRI can for instance be found in~\citep{Edupuganti2019}. An important aspect of this approach is that it relies on the known likelihood of observations.

A different approach was proposed in \cite{Adler2018DeepInversion} based on a conditional Wasserstein Generative Adversarial Network (cWGAN). In this formulation, a generative model is trained to sample high quality images conditioned on degraded observations. 
At test time, independent samples from the posterior distribution induced by the GAN are obtaind by simply sampling different latent space variables. Similar frameworks based on other conditional generative models have alse been proposed, such as~\citep{Denker2020} using conditional Normalizing Flows, or \citep{Tonolini2019} using conditional VAEs. 
While this conditional generative model approach allows for fast inference, thanks to ancestral sampling, it is worth noting that none of these models include an explicit \textit{data consistency} step, nor do they make use of a test-time likelihood. As a direct consequence, for instance, these models cannot be used on data with different noise levels as seen during training.

\section{Deep Denoising Score Matching for Posterior Inference}

Our main objective in this work is to perform probabilistic inference over a Bayesian model typically represented as 
\begin{equation}
    p(\xb | \yb) \propto p(\yb | \xb)  p(\xb)  \label{eq:bayes}
\end{equation}
where $\yb$ are some measurements and the posterior 
$p(\xb | \yb)$ is the distribution of possible solutions $\xb$ compatible with observations and $p(\xb)$ a prior knowledge. The problem-specific likelihood $p(\yb | \xb)$ encodes the forward process of the model and accounts for observational noise, while the prior $p(\xb)$ encapsulates any a priori information we have on the solution of the problem. Our goal for UQ is to sample solutions $\xb$ belonging to that posterior. Multiple inference techniques can be leveraged for sampling from this posterior $p(\xb | \yb)$, but for high-dimensional problems modern techniques rely on gradient-based methods, including Variational Inference~\citep{Hoffman2013} and Langevin Diffusion, or HMC~\citep{Neal2011}. All of these techniques have in common that they only require having access to the \textit{score} $\nabla_x \log p(x)$ of the target distribution. We illustrate in particular in \autoref{sec:score-hmc} how Metropolis Hastings-calibrated HMC can be conducted with only knowledge of the score.\\
Two terms will enter in the score of the posterior distribution in \autoref{eq:bayes}, the score of the likelihood, and the score of the prior. In many problems, such as in the MRI problem presented later in this work, the likelihood score can be derived analytically, only the score of the prior remains unknown but can be learned from data by score matching. 

\paragraph{Deep Denoising Score Matching}
As originally identified by \cite{Vincent2011} and \cite{Alain2013}, the score of a given target distribution $P$ can be modeled using a DAE, i.e. by introducing an auto-encoding function $\boldsymbol{r}: \mathbb{R}^{n} \times \mathbb{R} \mapsto \mathbb{R}^{n}$ trained to reconstruct under an $\ell_2$ loss a true $\xb \sim P$ given a noisy version $\xb' = \xb + \boldsymbol{n}$ with $\boldsymbol{n} \sim \mathcal{N}(0, \sigma^2 \boldsymbol{I})$. 
An optimal denoiser $\boldsymbol{r}^\star$ would then be achieved for:
\begin{equation}
    \boldsymbol{r}^\star(\xb', \sigma) = \xb' + \sigma^2 \nabla_{\xb} \log p_{\sigma^2}(\xb')
\end{equation}
where $p_{\sigma^2} = p \ast \mathcal{N}(0, \sigma^2)$. In other words, the optimal denoiser is closely related to the score we wish to learn and when the noise variance $\sigma^2$ tends to zero, should exactly match the score of the target density.  
In practice to learn this score efficiently, we adopt the residual noise-conditional denoising score matching technique proposed by \cite{Lim2020}, and train a model to minimize the following Denoising Score Matching loss:
\begin{equation}
    \mathcal{L}_{DSM} =  \underset{\xb \sim P}{\mathbb{E}} \underset{\begin{subarray}{c}
  \boldsymbol{u} \sim \mathcal{N}(0, I) \\
  \sigma_s \sim \mathcal{N}(0, s^2)
  \end{subarray}}{\mathbb{E}} \parallel \boldsymbol{u} + \sigma_s  \boldsymbol{r}_{\theta}(\xb + \sigma_s \boldsymbol{u}, \sigma_s) \parallel_2^2
    \label{eq:dsn}
\end{equation}
In this formulation, the network $\boldsymbol{r}_{\theta}(\xb, \sigma)$ is now directly modeling the score $\nabla_{\xb} \log p_{\sigma^2}(\xb)$ of the Gaussian-convolved target distribution.



\paragraph{Annealed Hamiltonian-Monte Carlo Sampling} Given the noise-conditional neural scores learned with the procedure described above, it is now possible to use a variety of inference methods to access the Bayesian posterior. In this work, we adopt an annealed HMC procedure which provides an efficient way to obtain parallel batches of independent samples from the target posterior despite the high dimensionality of the problem. This is a sampling procedure closely related to the Annealed Langevin Diffusion proposed in ~\citep{Song2019,Song2020}, but benefits from the faster Hamiltonian dynamics and Metropolis-Hastings calibration (see \autoref{sec:score-hmc}).

To build our procedure, we consider a Gaussian-convolved version of our target density:
\begin{equation}
    \label{eq:tempered-dens}
    \log p_{\sigma^2}(\xb | \yb) = \log p_{\sigma^2}(\yb |\xb) + \log p_{\sigma^2} (\xb) + cst
\end{equation}
where $\sigma^2$ plays the role of the inverse temperature found in classical annealing. The likelihood can be obtained analytically, and in the case of a Gaussian likelihood takes the following form: $\log p_{\sigma^2}(\yb |\xb) = -\frac{\parallel \xb - f(\xb) \parallel_2^2 }{2(\sigma_n^2 + \sigma^2)} + cst $, where $\sigma_n^2$ is the noise variance in the measurements. As for the prior term, the noise-conditional score network introduced above already models the score of $\log p_{\sigma^2}$. 
This distribution is gradually annealed to low temperatures and the chain progressively moves towards a point in the target distribution. 

\section{Application to Bayesian Inverse Problems}

\paragraph{The MRI Problem}
Magnetic Resonance Imaging (MRI) is a non-invasive modality used to probe soft tissues. 
Compressed sensing, introduced for MRI by~\cite{Lustig2007} is used to reduce its significant acquisition time, and recently, deep learning approaches(see for example~\citep{Schlemper2018,Hammernik2018,Pezzotti2020}) have been shown to perform extremely well on the reconstruction problem.
The idealized reconstruction problem is usually formalized in the following way, for the single coil setting, with uniform acquisition:
\begin{equation}
    \label{eq:inv_prob}
    \yb = M_{\Omega} \Fb \xb + \boldsymbol{n}
\end{equation}
where $\yb$ is the acquired Fourier coefficients, also called the k-space data, $M_{\Omega}$ is a mask, $\Fb$ is the classical 2D Fourier transform, $\xb$ is the real anatomical image, and $\boldsymbol{n} \sim \mathcal{N}(0, \sigma_{n}^2)$ is measurement noise (we set  $\sigma_{n} = 0.1$ in our experiments). 
The data we consider is the single coil data with the Proton-Density (PD) contrast from the fastMRI dataset introduced in~\citep{Zbontar}. 
For comparison with deep learning approaches, we use the state-of-the-art Primal-Dual net enhanced with a U-net for image correction (UPDNet) from~\citep{Ramzi2020}. 
The under-sampling was done retrospectively using an acceleration factor of $4$ and a random mask as described for knee images in~\citep{Zbontar}.

\paragraph{Results}

We use a simple residual U-net inspired by~\cite{Ronneberger} with ResNet building blocks from~\citep{He2016a} as our score network. In order to promote the regularity of the learned scores, we regularize the spectral norm of each convolutional layer, and we find that setting the spectral norm to $\simeq 2$ yields the best results. More details about the network architecture and training can be found in appendix~\ref{sec:mri}. 
We then sample from the posterior following our annealed HMC procedure down to relatively low temperatures, and apply one final denoising step on the last sample from the chain, following the Expected Denoised Sample (EDS) scheme detailed in~\citep{Jolicoeur-Martineau2020}. 
\begin{figure}
\centering
\includegraphics[width=\textwidth]{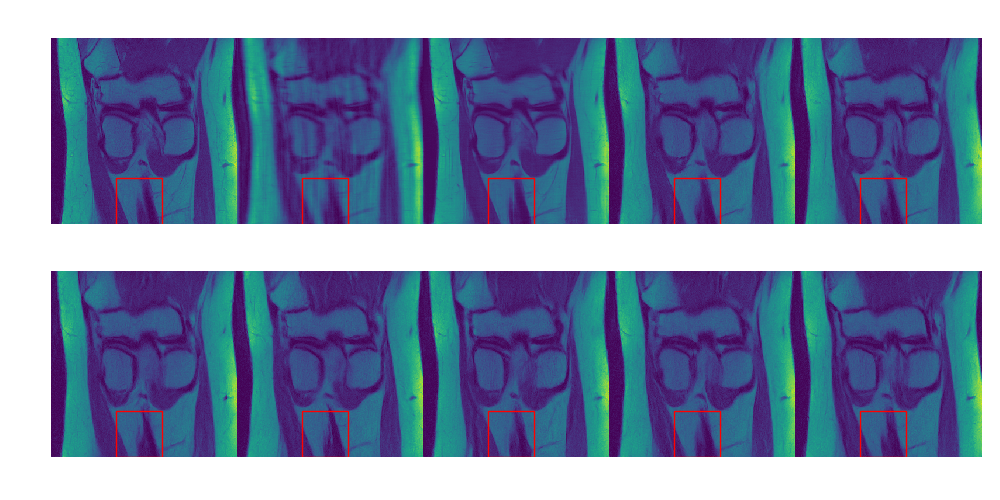}
\caption{\small
\label{fig:mri-recon} Bayesian posterior sampling for MRI reconstruction. The top leftmost image is the ground truth image. The top second to the left image is the zero-filled retrospectively under-sampled image $\Fb^T \yb$. The top third to the left image is the reconstruction of the under-sampled image by the UPDNet. All the other images are denoised samples from the estimated posterior distribution obtained by a tempered HMC. The zero-filling achieves a PSNR of 25.55 dB, each sample 27.63 dB on average, the mean of the samples 30.04 dB and the neural network 32.15 dB. A zoom of the region in the red square is provided in appendix~\ref{subsec:mri-zoom}. An animation of posterior samples is available online \imgur .}
\vspace{-0.5cm}
\end{figure}
Figure~\ref{fig:mri-recon} compares samples from the MRI posterior to the ground truth (top left), zero-filled image (top second left), and UPDNet reconstruction (top center). We see that although individual samples carry slightly less details than the neural network reconstruction, confidence in any particular part of the image can be gauged by looking for variability or stability across multiple independent posterior samples. 
We highlight in particular the red region, where posterior samples show significant variability, indicating that this part of the image is poorly constrained by data. 
In contrast, a direct neural network reconstruction (top center) does not match the ground truth in that region, and does not provide an estimate of uncertainty which may lead a physician to misinterpret the image.

\section{Conclusions and Discussions}
We have presented in this work the first (to the best of our knowledge) instance of a framework for Bayesian inverse problems based on Deep Denoising Score Matching and applied to MRI reconstruction and Uncertainty Quantification. We illustrated the merits of this approach on an MRI example where the ability to sample from the full posterior highlights what features present in a reconstruction are not actually constrained by data.\\
We also showed that the Deep Denoising Score Matching approach can effortlessly be used for dimensions as high as $320 \times 320 \times 2$. 
It is therefore much more scalable than GAN- or VAE-based models.
We note however that finding an optimal HMC annealing schedule and temperature-adaptive step-size proved difficult. 
In general we did not manage to lower the temperature below a certain level, prompting us to resort to a denoising step on the last samples from our HMC chains. This is a direction for further research.


\begin{ack}
We acknowledge the financial support of the Cross-Disciplinary Program on Numerical Simulation of CEA for the project entitled SILICOSMIC.
We also acknowledge the French Institute of development and ressources in scientific computing~(IDRIS) for their AI program allowing us to use the Jean Zay supercomputer's GPU partitions.
\end{ack}


\medskip


\bibliography{references.bib}
\bibliographystyle{unsrtnat}

\appendix
\section{MRI experiments}\label{sec:mri}
\subsection{Network architecture} \label{subsec:mri-network}
The network is a 3-scale U-net with residual blocks composed of 3 convolutions followed by a batch normalisation. 
Each batch normalisation is followed by a ReLU non-linearity except the last one.
A projection is used for the input of residual blocks whose number of input channels is not the same as that of output channels.
Each scale uses 2 residual blocks for the downsampling path and 2 residual blocks for the upsampling path.
Downsampling is performed via average pooling and upsampling is performed via up-convolution (as designed in~\citep{Ronneberger}), in order to avoid checkerboard artefacts.
We use the following sequence of number of channels: [32, 64, 128].
In order to deal with MRI images, that are complex in nature, we concatenate the real and imaginary part of the image at the input of the network, forming effectively a 2-channel $320\times320$ image.
For input noise level conditioning, we concatenate a noise standard deviation map to the input and in the lower scale of the network.
Following the recommendation of~\cite{Song2020}, we also divide the output of the network by the absolute value of the noise power.
Finally, we use Spectral Norm regularisation from~\citep{Yoshida2017} in order to make sure the score does not take inconsistent values in low-density regions. 
The Spectral Norm regularisation indeed forces the spectral norm (maximum eigenvalue) of each layer to a certain value which in turn lowers the Lipschitz constant of the network, preventing it to blow in unseen regions.
The influence of the Spectral Norm has been studied in the case of Plug-and-Play approaches by~\cite{Ryu2019Plug-and-playDenoisers}.

\subsection{Network training} \label{subsec:mri-training}
We use the Adam optimizer for network training, with a learning rate of $10^{-4}$.
We add a white Gaussian noise of variance $\sigma_s^2$ to the images scaled by a factor of $10^6$.
$\sigma_s$ is drawn on-the-fly from a Gaussian distribution of variance $s=50$.
This means that at training time, the standard deviation of the noise can be negative, following the recommendation of~\cite{Lim2020} to go from extrapolation to interpolation.

\subsection{HMC procedure} \label{subsec:mri-hmc}
The HMC procedure starts from a zero-filled reconstruction of the image with added white Gaussian noise of variance $\sigma_{init} = 100$.
For the reduction of noise standard deviation at reduction step, we use a factor $\gamma = 0.995$, following the recommendation of~\cite{Song2020}.
We take a step size $\alpha$ dependent of the sampling temperature at step $i$,  $\alpha = \epsilon \left( \frac{\sigma_i}{\sigma_0}\right)^{1.5}$ with $\epsilon=10$.

\subsection{Zoom on Reconstruction} \label{subsec:mri-zoom}
\begin{figure}
\centering
\includegraphics[width=\textwidth]{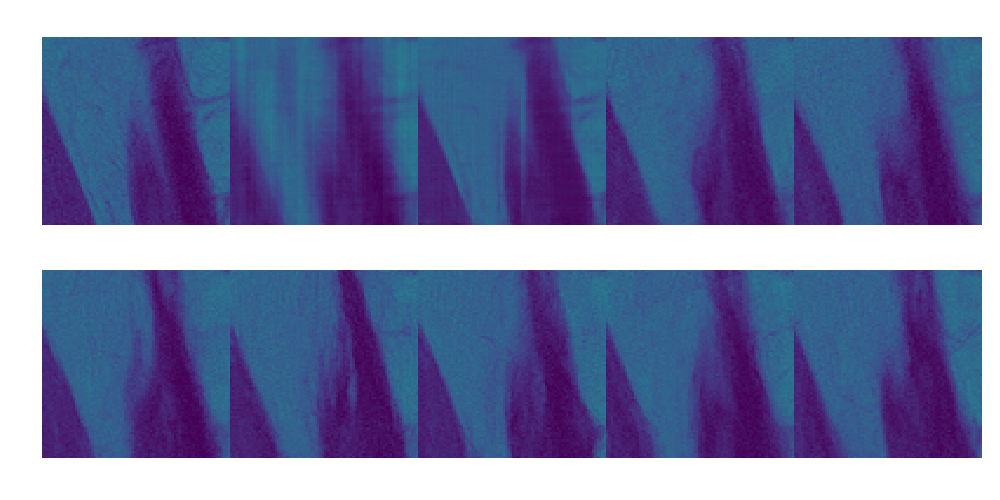}
\caption{\label{fig:mri-recon-zoom}
Zoom of Bayesian posterior sampling for MRI reconstruction. The ordering is the same as in Figure~\ref{fig:mri-recon}.}
\end{figure}

The zoom on this part of the reconstruction shown in Figure~\ref{fig:mri-recon-zoom}, illustrates a failure of the Neural network reconstruction.
It indeed produces a sharp artifact that might hinder the examiner's conclusions.
The bayesian sampling allows us to see that this region is not necessarily very sharp.

\section{Score-Based Hamiltonian Monte Carlo}
\label{sec:score-hmc}
We used HMC for sampling from the posterior distribution. The sampling procedure follows the leapfrog integrator presented bellow:

\begin{equation}
    \begin{array}{lcl} \boldsymbol{m}_{t+\frac{\alpha}{2}} & = & \boldsymbol{m}_t + \dfrac{\alpha}{2} \nabla_{\xb} \log p(\xb_t | \yb) \\ 
    \xb_{t+\alpha} & = & \xb_t + \alpha \boldsymbol{\mathrm{M}}^{-1} \boldsymbol{m}_{t+\frac{\alpha}{2}} \\
    \boldsymbol{m}_{t+\alpha} & = & \boldsymbol{m}_{t+\frac{\alpha}{2}} + \dfrac{\alpha}{2} \nabla_{\xb} \log p(\xb_{t+\alpha} | \yb)
    \end{array}
    \label{eq:hmc}
\end{equation}

where $\alpha$ is the step size, $\boldsymbol{m}$ the momentum and $\boldsymbol{\mathrm{M}}$ is a preconditioning matrix taking into account the space metric, but in our case the identity matrix. The gradient of the log posterior is computed as $\nabla_{\xb} \log p(\xb_t | \yb) = \nabla_{\xb} \log p(\yb | \xb_t) + \nabla_{\xb} \log p(\yb | \xb_t)$, using the inverse Bayes theorem in equation \ref{eq:bayes}.

It is worth noting that the discretized HMC with final step size $\epsilon$ is not guaranteed to converge to the desired target distribution. This is the motivation for the Metropolis-Hastings algorithm~\citep{Metropolis1953, Hastings1970}, which corrects for the discretization errors with a correction step. Concretely, a new step is proposed using \autoref{eq:hmc}, but it is accepted with probability of the form $\min\{1 ,  p(\xb^*) q( \xb_n | \xb^* )/ p(\xb_n) q(\xb^* | \xb_n) \}$ where $\xb_n$ is the last accepted point in the chain, $\xb^*$ is the proposed update, $p$ is the target density, and $q$ is the proposal corresponding to taking a step of \autoref{eq:hmc}. In the context of a score-only approach, we do not have access to $p$ and therefore cannot compute directly this acceptance probability. We can nonetheless estimate from the scores themselves by using a path integral: 
\begin{equation}
\log p(\xb_*) - \log p(\xb_n) =  \int_0^1 \nabla_{\xb} \log p( t * (\xb_* - \xb_n) + \xb_n) \cdot (\xb_* - \xb_n) dt
\end{equation}
This integral can be evaluated numerically to arbitrary precision (at the cost of additional score evaluations), but we find excellent agreement with the theoretical value on toy model densities like the two-moons distribution with a simple 4 points Simpson integration rule.

This illustrates that for gradient-based MCMC samplers, which also includes Metropolis Adjusted Langevin Algorithm, these algorithms can fully be implemented with only knowledge of the score.

\end{document}